\documentclass{elsarticle}
\usepackage{epsfig} 
\usepackage{amsmath}
\usepackage{bm}
\usepackage{graphicx}
\usepackage{multirow}
\usepackage{booktabs}
 \usepackage{graphicx}
\usepackage{hyperref}

\journal{Journal of \LaTeX\ Templates}

\begin{document}

\begin{frontmatter}

\title{Inversion of biological strategies in engineering technology: in case underwater soft robot}

\author[mymainaddress]{Siqing Chen}
\ead[url]{}

\author[mymainaddress]{He Xu\corref{mycorrespondingauthor}}
\cortext[mycorrespondingauthor]{Corresponding author}
\ead{railway_dragon@sohu.com}

\author[mymainaddress]{Xueyu Zhang}
\ead[url]{}

\author[mymainaddress]{Zhen Ma}
\ead[url]{}

\address[mymainaddress]{Department of Mechatronics Engineering, Harbin engineering university, Harbin, China}

\begin{abstract}
  This paper proposes a biomimetic design framework based on biological strategy inversion, aiming to systematically map solutions evolved in nature to the engineering field. By constructing a "Function-Behavior-Feature-Environment" (F-B-Cs in E) knowledge model, combined with natural language processing (NLP) and multi-criteria decision-making methods, it achieves efficient conversion from biological strategies to engineering solutions. Using underwater soft robot design as a case study, the effectiveness of the framework in optimizing drive mechanisms, power distribution, and motion pattern design is verified. This research provides scalable methodological support for interdisciplinary biomimetic innovation.
\end{abstract}


\end{frontmatter}


\section{Introduction}
The core process of biomimetic inspired design can be divided into four progressive stages: problem definition, biological prototype screening, principle extraction, and engineering technology transformation\cite{RN259}. This paradigm is essentially a cross-domain knowledge reconstruction process, utilizing existing biological characteristics, behaviors, and functions to correspond to features, behaviors, and similar functions in engineering, with the key being the efficiency of knowledge mapping between biological systems and engineering systems\cite{RN260}.

The cognitive bottleneck in current research areas lies in the fact that the high complexity of biological systems often makes it difficult to pinpoint key strategic information, while the existing knowledge framework of engineering systems struggles to effectively integrate with biological strategy knowledge. The biological knowledge system is the foundation of biomimetic design, and solving engineering problems involves analogizing biological strategies\cite{RN261}. Researchers with a biological background can explain the operational rules of natural systems well but lack knowledge reserves for engineering problems\cite{RN262}. Engineers working in this field commonly encounter systemic barriers in identifying biological strategies, constrained by the professional barriers of the biological terminology system and the technical limitations of interdisciplinary knowledge expression\cite{RN262}\cite{RN261}. Therefore, constructing an intelligent matching mechanism between biological characteristics and engineering parameters, and improving the technical processes for screening biological prototypes and converting engineering technologies, are important research directions for enhancing the effectiveness of biomimetic design.

In biomimetic design, the key steps lie in two points: 1) finding biological strategies corresponding to engineering problems (biological strategy mapping); 2) expressing biological characteristics and behaviors in an engineering form (biological strategy inversion). Biological strategy mapping is the foundation of biomimetic design, and it is crucial. The process involves systematically deconstructing the functional advantages formed by organisms during evolution using biological methods, establishing a cross-domain mapping relationship between biological systems and engineering requirements, and applying reverse engineering principles to convert biological strategies into implementable technical solutions. Current academic achievements focus on the digital representation of biological knowledge and morphological imitation, with insufficient attention paid to the integration mechanisms at the level of complex systems. Biomimetic engineering design falls within the realm of fuzzy decision-making scenarios, where descriptions of biological strategies mostly belong to natural language corpora\cite{RN263}.

Existing bionic design research often relies on experience-driven approaches, lacking a systematic knowledge framework and automated tools. This paper proposes a bionic design framework that integrates knowledge modeling, text analysis, and decision optimization. The core contributions include: F-B-Cs in E Knowledge Model: standardizing the description of biological strategy functions (Function), behaviors (Behavior), characteristics (Characteristic), and environmental associations (Environment). Text-driven strategy transformation: multi-label classification based on GPT models and engineering knowledge base corrections to achieve automated mapping of biological strategies. Hybrid multi-criteria decision-making method: combining VIKOR with rank correlation analysis to balance designer preferences and objective metrics, selecting the optimal engineering strategy.


\section{Related work}
In the process of bionic strategy inversion, solution generation based on functional model and solution generation based on conflict resolution are two mainstream solutions from biological prototype to engineering implementation \cite{RN264}. These two dimensions correspond to the functional decomposition of biological system and the reconciliation of contradictions respectively.

Nagel et al. \cite{RN265}proposed a biologically inspired dual-drive design framework, establishing the theoretical foundation for cross-domain knowledge transfer through problem-driven (engineering problems-biological solutions mapping) and solution-driven (new biological discoveries-engineering migration) bidirectional pathways. Hancock et al. \cite{RN266}introduced an evolution-driven design method, abstracting biological characteristics into "product genes," and simulating biological evolution through structural-function mapping and genetic algorithms to achieve autonomous optimization of components in biomimetic composite materials. Wiltgen et al. \cite{RN267}, on the other hand, introduced Case-Based Reasoning (CBR) technology, building a biological knowledge base that includes cases such as bat sonar and spider silk mechanics, using feature matching algorithms to shorten the generation cycle of engineering solutions.

Conflict resolution-based solution generation focuses on the path of resolving contradictions in bio-functional engineering, which can be summarized as a process of "conflict identification-cross-domain modeling-intelligent resolution-functional validation." Currently, TRIZ contradiction matrices or quality function deployment identify technical conflicts between biological characteristics and engineering requirements, and further operations are carried out using corresponding solutions from the contradiction matrix \cite{RN268}. Similar to these methods, Zhao et al. \cite{RN269} addressed the technical conflict between noise reduction and aerodynamic performance in the trailing edge serrated structure of wind turbine blades by developing a multi-disciplinary optimization-based aerodynamic-structural collaborative design method, creating an optimized design platform that balances noise reduction and aerodynamic efficiency. This ultimately resulted in a serrated airfoil with a 1.9\% increase in lift-to-drag ratio, a 32.5\% increase in lift coefficient, and reduced noise; Zuo et al.\cite{RN270}tackled the physical conflict of flexible AC electrochromic devices struggling to achieve real-time wide-range color tunability and mechanical stability by developing a double-layer stacked luminescent structure based on dielectric difference regulation, optimizing the dielectric matrix material to enhance the device's mechanical robustness and simulate biological camouflage and visual communication functions. The essence of this process is to achieve a transformation from contradictory opposition to collaborative innovation through the balance of biodiversity advantages and engineering constraints.

\section{Inversion method}
\subsection{Inversion of biological strategies in engineering technology}
Biological distribution strategies represent solutions to biological challenges that have evolved through natural selection in nature. Biomimetic design, on the other hand, involves mapping engineering problems into the realm of nature, drawing upon and adopting similar strategies to solve these engineering issues. This process of borrowing is known as the reverse problem of biological strategies in engineering technology. The screening process for biological prototypes merely identifies and extracts numerous potential strategies; whether and how these strategies can be effectively applied to engineering practice still requires further exploration and validation. It is worth noting that these strategies derived from biology are essentially aimed at problems within biological systems, whereas engineering problems often have different complexities and constraints, thus inherently differing. Given this, exploring how to effectively convert biological strategies into engineering strategies, evaluating their applicability in the engineering field, and selecting the optimal strategies constitutes another core topic in biomimetic design research.

\subsubsection{A textual knowledge framework of biological strategies and engineering technologies
}
\begin{figure*}[htbp]
  \centering
  \includegraphics[width=12cm]{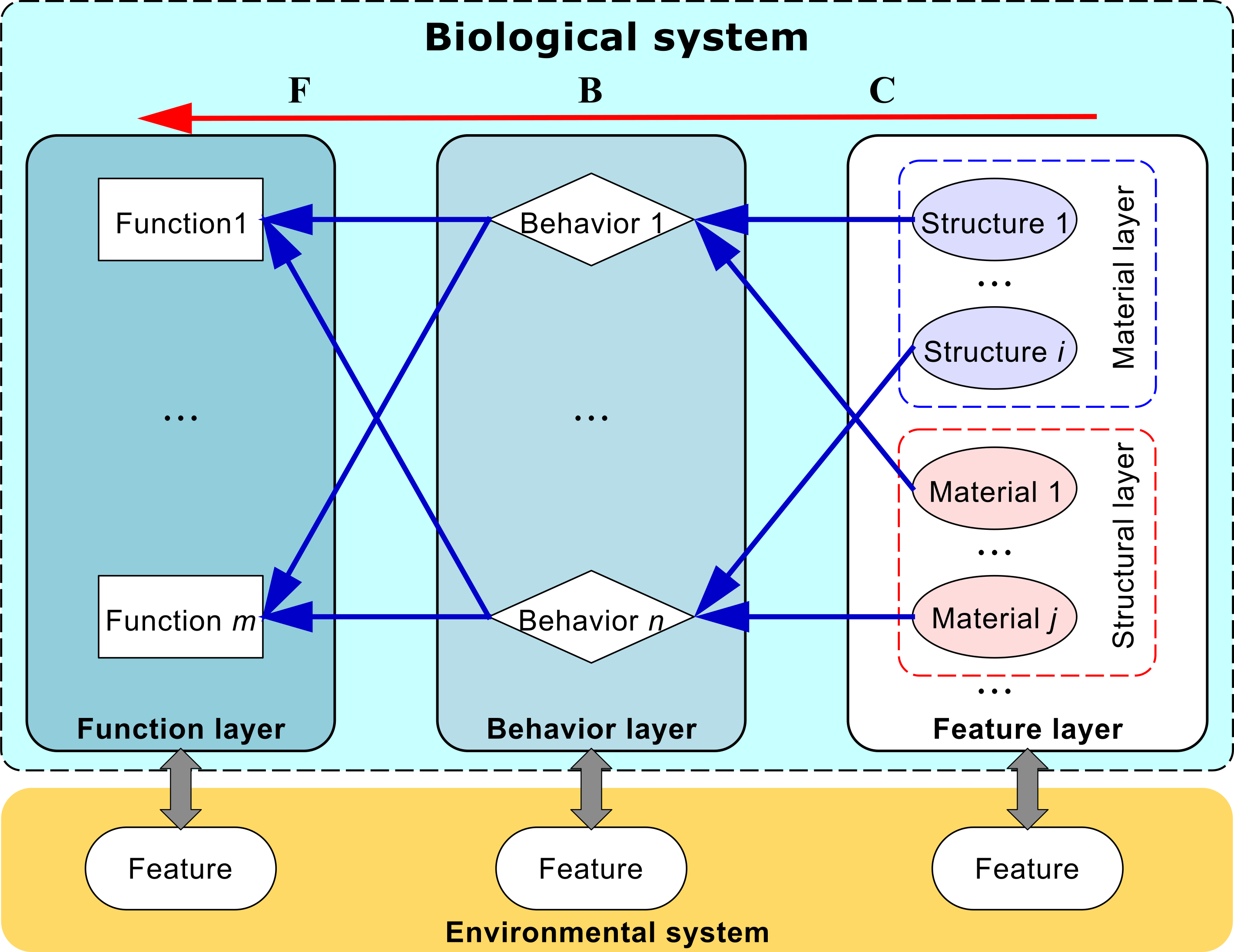}\\
  \caption{Knowledge framework based on characteristics-behaviors-functions in environmental systems
  }
  \label{fig_kf}
\end{figure*}

This study constructs a knowledge framework system aiming at connecting engineering and biology, which is used to explain the causal chain of example behaviors in a specific environmental background. The specific structure of the model framework is shown in Fig. \ref{fig_kf}.

The biological system of this model comprises three levels: function (Function, F), behavior (Behavior, B), and characteristics (Characteristic, C). Driven by environmental properties, biological strategies form corresponding material and structural characteristics. These characteristics constitute the fundamental elements for coping with environmental constraints and achieve the associated functions. Under certain conditions, the functions, behaviors, and characteristics within these strategies also influence the current environment, further altering its features to adapt to their own functions and behaviors.

Function representation describes the purpose of biological systems and serves as the foundational framework for organizing and indexing biological strategies and knowledge analogies. Methods of function identification include action description, flow transformation (energy flow, material flow, and signal flow), and state transition. Action description involves combining verbs (entity actions) and nouns (action objects) to describe functions, where verbs (functional predicates) represent the system's actions, and nouns indicate the action objects. Flow transformation is used to describe the conversion between input and output flows in system functions, involving fixed actions ("transforming" verbs) and two nouns (input/output objects). State transition represents the function of a system by describing the change from an initial state to a final state, focusing on changes in state parameters, including one noun (the object being transformed) and one verb (the way the state changes is described).

Behavioral representation is used to describe the operational mechanisms of a system, representing the process of system function realization. In specific working environments, the behavioral modules of the system will exhibit specific behavior processes in response to given drive inputs. The causal relationships in behavioral representations describe the mechanisms of system function realization and the interactions between functions, behaviors, characteristics, and working environments. System behavioral knowledge representation involves explaining the state transitions and their causes of the system. It is worth noting that the same feature can achieve different functions through different behavioral mechanisms; the same function can also be realized through different behavioral paths, corresponding to different characteristic modules. Its description system includes temporal processes and causal relationships. Temporal processes are the ordered arrangement of functions of different objects. Causal relationships consist of a temporal process description (cause description), a single object's function description (effect description), and a conjunction indicating the causal relationship.

Feature representation serves as the carrier of functions, used to describe the composition of a system. Environmental representation is the sum of external factors that influence system operation but are not part of the system instance. Therefore, in this paper, all bionic design texts and knowledge will be designed as a model path (F-B-Cs model in E) that combines single functions, single behaviors, and multiple characteristics under specific environments. The causal relationship of behavior is used to explain the reasons and purposes behind the behavior; during text inversion, this information can often be ignored. However, the sequential processes of multiple behaviors can be handled concurrently in a synchronous process. Thus, from a scientific perspective, the "F-B-Cs model in E" model path is reasonable and feasible.

\subsubsection{The transformation process of biomimetic features
}
In the process of converting biological strategy texts into engineering strategy frameworks, it is necessary to construct corresponding biological knowledge based on the aforementioned knowledge expression framework. This biological knowledge aims to clearly define functions, behaviors, characteristics, and environmental modules in biological strategies. By transforming biological strategy texts into a standardized structure, not only does it help designers systematically organize their design ideas, but it also further enhances the accuracy of language processing dominated by commercial models. Therefore, the first step in biomimetic feature transformation is to systematically classify the originally scattered biological strategy texts into four dimensions: function, behavior, characteristic, and environment, and output them in a standardized format.

\begin{figure}[htbp]
  \centering
  \includegraphics[width=12cm]{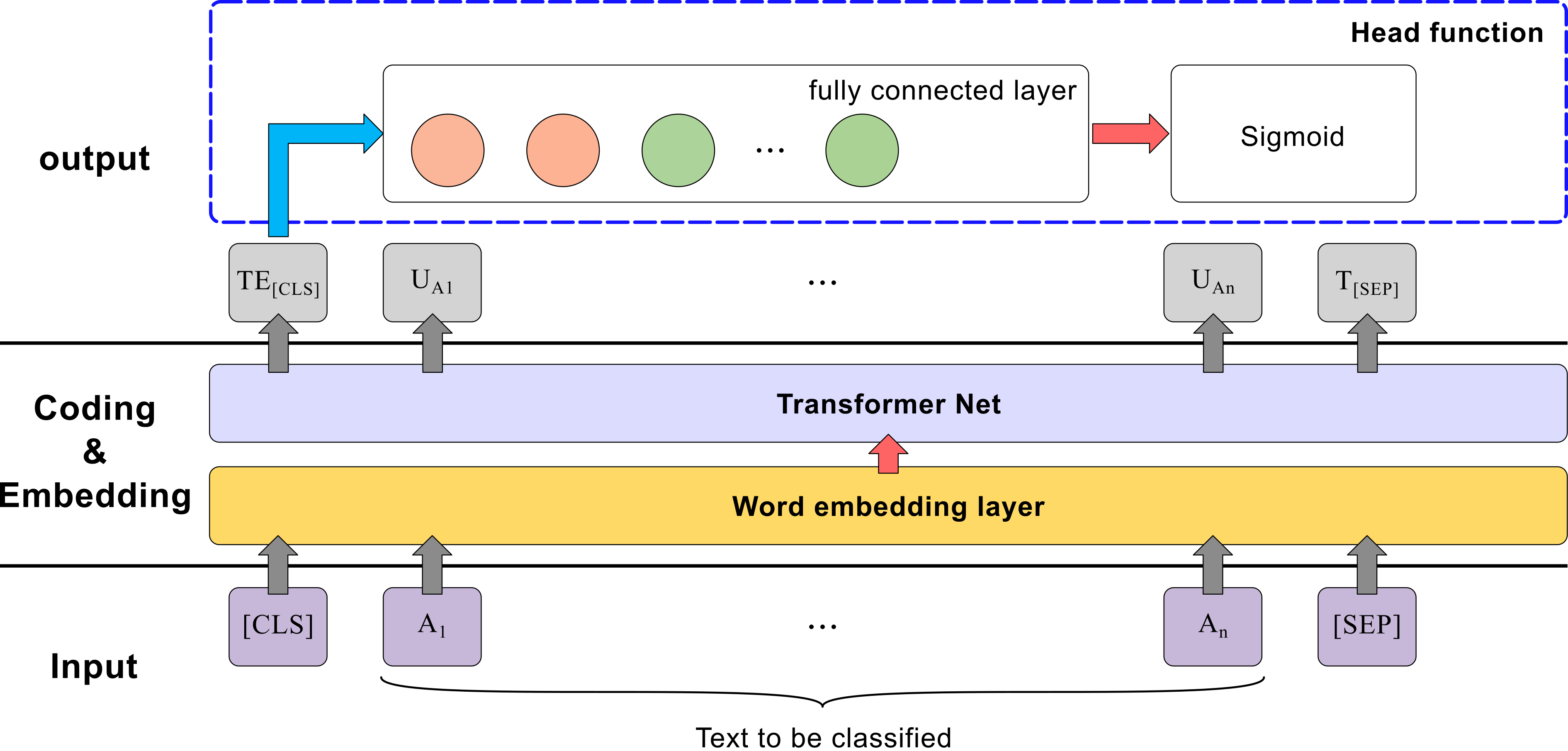}\\
  \caption{Structure of the label classification model.}
  \label{fig_tranf}
\end{figure}

The previously used NLP model is also applicable to text label classification tasks. Given the complexity and diversity of language, a single sentence often contains information from multiple aspects. Therefore, in this context, the language processing task should be reasonably set as a multi-label classification task, allowing a sentence to belong to multiple labels simultaneously. When using the GPT model to calculate probabilities for each label, the text was appropriately adjusted based on the original GPT model architecture: the input part was set to a single sentence, while the output part added a fully connected layer and a classification layer to achieve precise label assignment functionality. The structure of the model is shown in Fig. \ref{fig_tranf}. Since the labels in the label classification task are not mutually exclusive, the Sigmoid function was chosen as the final classification layer function.

The basic data for text label classification is consistent with the previous section. After the same data cleaning and preprocessing, sentences from different texts are extracted based on sentence symbols as training samples. The 18,888 bionic statements screened out from relevant bionic documents will first be fed into the LLM, where the model will act as a biologist and engineer to generate initial labels. Then, the statements and labels will be organized into sets, with 100 data points per set, and 3\% of the samples will be randomly selected for manual review. This process will continue until all statements and labels have been reviewed by experts and no more unqualified items can be identified.

Due to the richness and complexity of language, the text content undergoes certain data enhancement. Sentences are randomly replaced with LLM in terms of language style at a certain ratio, and combined with post-review tags to form samples. Negative samples do not need to be created for the label classification task. The sample generation algorithm is shown in Fig. kk.

In the sample design, the number of samples is controlled to 10,000. In the composition of the sample, the ratio between real samples and enhanced samples is 8:2 to ensure the balance and pertinence of the sample set.

With the assistance of GPT models, this study will label each sentence in the corpus for a single selected bionic strategy. These labels cover four dimensions: function, behavior, characteristics, and environment. It is worth noting that a single piece of text may have multiple such labels. Subsequently, based on these labels, bionic texts will be categorized into four classes from a longer text. With the help of commercial LLMs,these categories will be systematically summarized and analyzed according to the biological knowledge structure framework mentioned earlier.

The single text selected from the biological strategy library follows a specific path to construct the "F-B-Cs model in E" model path. This model comprises four core modules, all expressed in a unified generalized format: functions described in gerund form (Function), behaviors described by a sequence of gerunds (Behavior), features described using noun phrases with attributive (Characteristics), and (optional) environments described using noun phrases with attributives (Environment). Given the standardization requirements of the knowledge model, nouns involved in the knowledge formation stage can be directly replaced with corresponding engineering nouns according to the specific requirements of the text for engineering needs. This process facilitates the effective transformation of biological model knowledge into engineering solutions within the F-B-Cs model in E framework, marking the initial transition phase from biological strategies to engineering applications.

\begin{figure}[htbp]
  \centering
  \includegraphics[width=12cm]{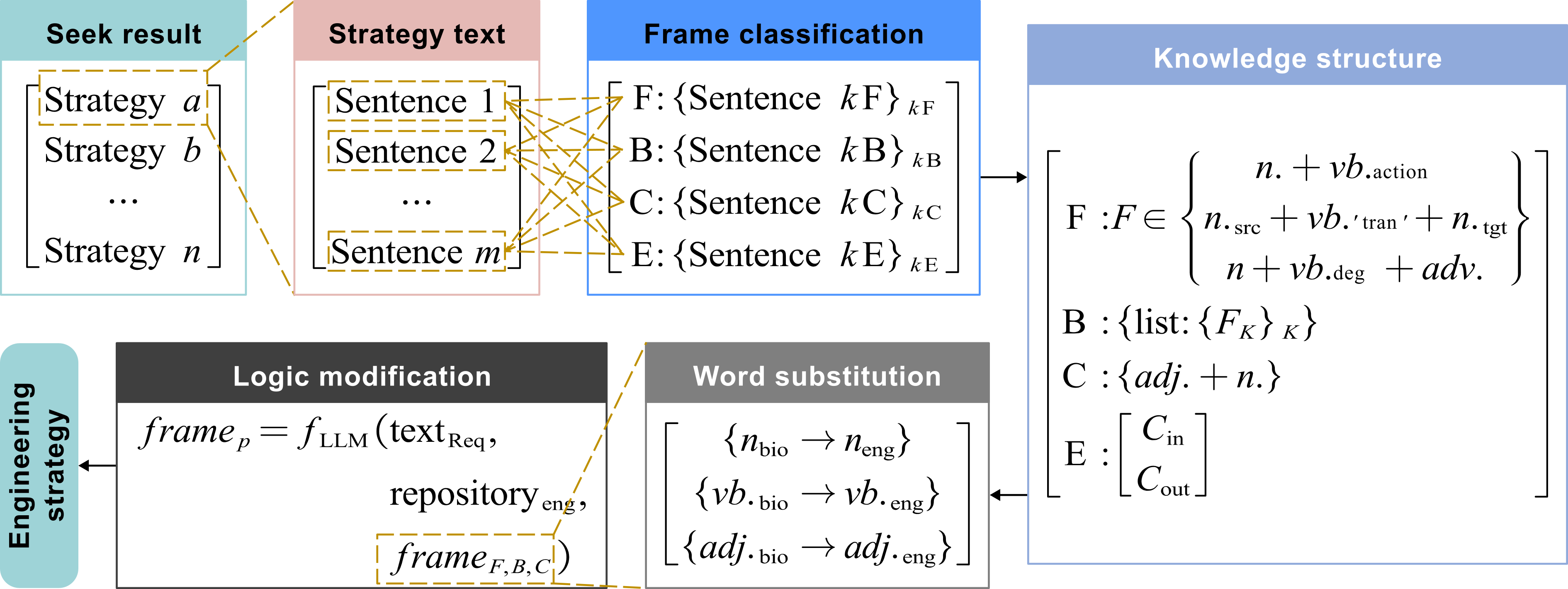}\\
  \caption{Bionic text conversion based on F-B-Cs model in E knowledge framework.}
  \label{fig_td}
\end{figure}

In this initial phase, the engineering model may have logical flaws, and the modifiers of phrases may not fully align with engineering standards. However, LLM can leverage its pre-prepared engineering knowledge base to make logical corrections to the forced correspondences. The correction process ensures that the engineering solutions remain consistent with the standard solutions in the knowledge base, meeting the objective theoretical logic of designers, thus facilitating the initial transition of biological strategies into engineering applications. This process is illustrated in Fig. \ref{fig_td}.

\subsubsection{The process of determining engineering strategies based on hybrid multi-standard decision}
Decision-making methods typically involve mapping judgment matrices that encompass all evaluation criteria. Many algorithms require constructing judgment matrices that meet consistency conditions, which not only increases the workload for designers but also raises scientific questions about the consistency testing standards of these matrices in academic circles. In contrast, rank correlation analysis does not require consistency testing and can effectively reduce the number of expert judgments needed. Therefore, the VIKOR method, as a multi-criteria decision-making approach based on ideal points, demonstrates significant advantages in establishing the connection between the ideal solution and statistical analysis. Due to its broad applicability, it can also be combined with rank correlation analysis in solving multi-criteria decision problems. Given this, this paper chooses to use the VIKOR method for comprehensive evaluation of strategies.

\subsubsection{The general process of the inversion of biological principles}
\begin{figure}[htbp]
  \centering
  \includegraphics[width=12cm]{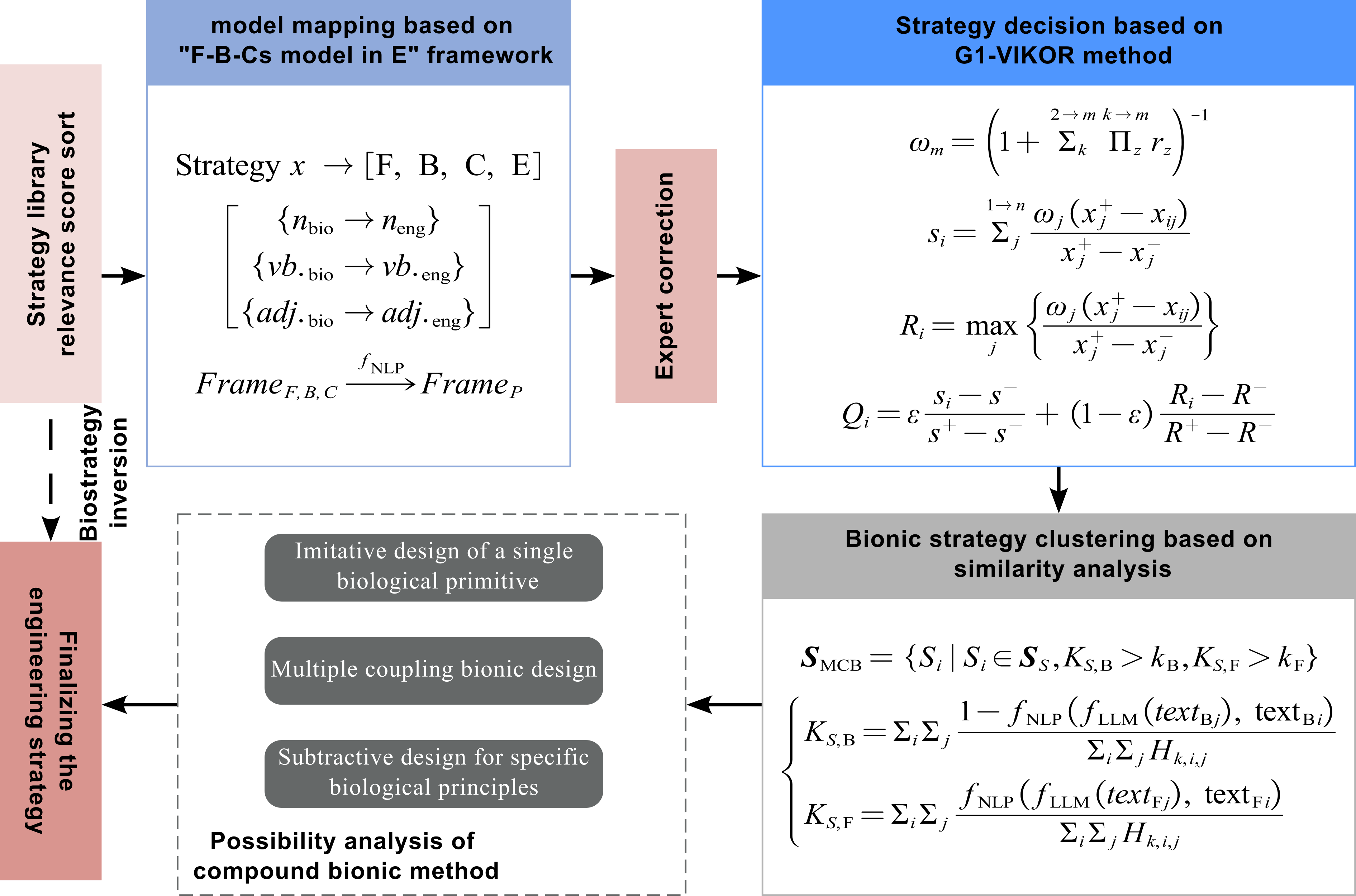}\\
  \caption{The total inversion process of biological strategy in engineering technology.}
  \label{fig_im}
\end{figure}

Due to the diversity and complexity of biological strategies, solving the inversion problem of biological strategies still relies on the final judgment of designers as experts in the field. In summary, this issue involves three aspects: reshaping text structure, updating text objects, and human verification by experts. The entire mapping process is shown in Fig. \ref{fig_im}, which outlines the overall workflow for the inversion of biological strategies in engineering technology.

First, the GPT model is used for text classification to populate the knowledge framework of the composite general knowledge strategy "F-B-Cs model in E," thereby facilitating the summarization of the knowledge framework by commercial LLMs.The LLM will leverage the part-of-speech characteristics of the knowledge framework, replacing the subject of biological strategies with that of engineering texts, while referencing the engineering logic in the knowledge base to refine the text within the framework, thus forming an initial mapping result for biological strategies. Subsequently, designers intervene to preliminarily screen all mapped strategies, removing those that do not fully align with engineering logic, and use the G1-VIKOR method to rank the remaining strategies. Designers can specify the priority of evaluation metrics for each strategy and sort them according to decision criteria. Then, designers cluster the top-ranked strategies based on similarity criteria and assess the likelihood of converting strategies into composite bionic methods based on content relevance. The final output is a set of executable engineering strategies that have been screened, ranked, and optimized.

In this paper, the optimal strategy selected is mapped to the transformation task in engineering applications, which is decomposed into the classification and word class conversion of text, and the ranking problem of multi-standard hybrid decision.

\subsection{Design process of underwater soft robot based on bionic inspiration}
As shown in the formula, the design object is systematically divided into three levels, aimed at addressing technical requirements in stages. These three levels are the system level, subsystem level, and component level. In each design level, there are two major components: technical requirement elements and technical processing elements. Faced with an unprocessed design object, designers, based on their professional background and knowledge, break down complex technical issues into single or multiple levels for targeted handling.

\begin{figure*}[htbp]
  \centering
  \includegraphics[width=12cm]{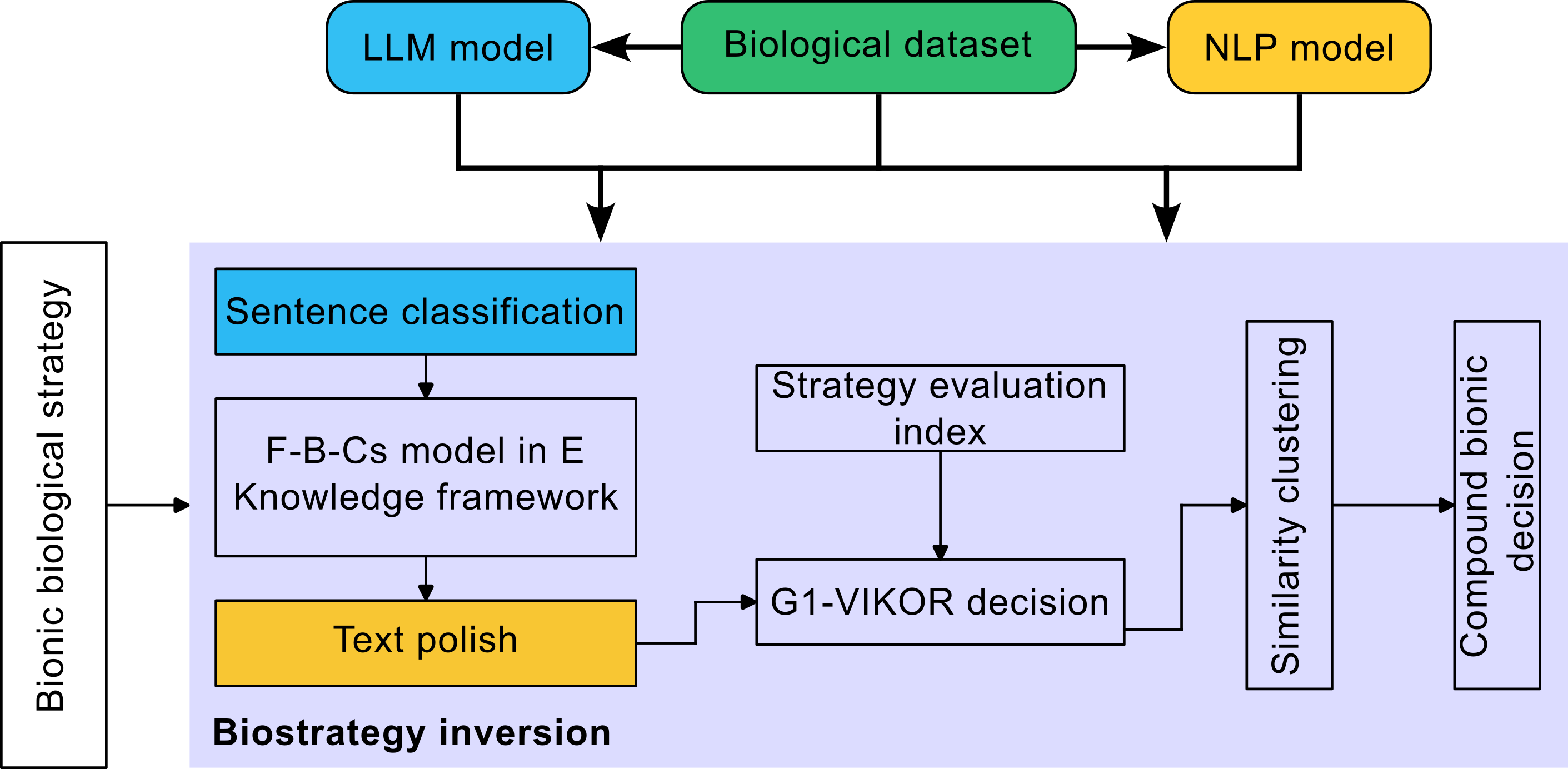}\\
  \caption{Bionic design process based on engineering mapping and strategy inversion.}
  \label{fig_ima}
\end{figure*}

The entire bionic design process is illustrated in Fig. \ref{fig_ima}, which shows the bionic design process based on engineering mapping and strategy inversion. Based on the concept of bionic-inspired design, this paper has developed a series of preprocessing procedures for issues related to the field of underwater soft robotics engineering. The core object of this design process is pressure-actuated underwater soft robots, aiming to explore their movement mechanisms in aquatic environments.

According to the existing literature and current engineering practice, pressure-driven soft robots face three challenges in their design and implementation: first, the uncertainty of the drive mechanism and the manufacturing complexity of fluid actuation components; second, the complexity of power generation and the instability of pressure supply; third, the diverse requirements for robot motion modes.

In response to the aforementioned three major challenges, this paper employs a biomimetic design strategy to address each one. Specifically, improvements to the drive mechanism (particularly the braking system) are categorized at the component level; optimization of power distribution and generation (mainly involving the pressure regulation system) is considered as a subsystem-level issue; and the design of the robot's motion strategy is elevated to the system level for consideration. Through this layered approach, the preliminary processing work for the design of the biomimetic underwater soft robot is completed.


\section{Case study}
\subsection{Soft underwater robot based on tail swing propulsion strategy}

\begin{figure*}[htbp]
  \centering
  \includegraphics[width=12cm]{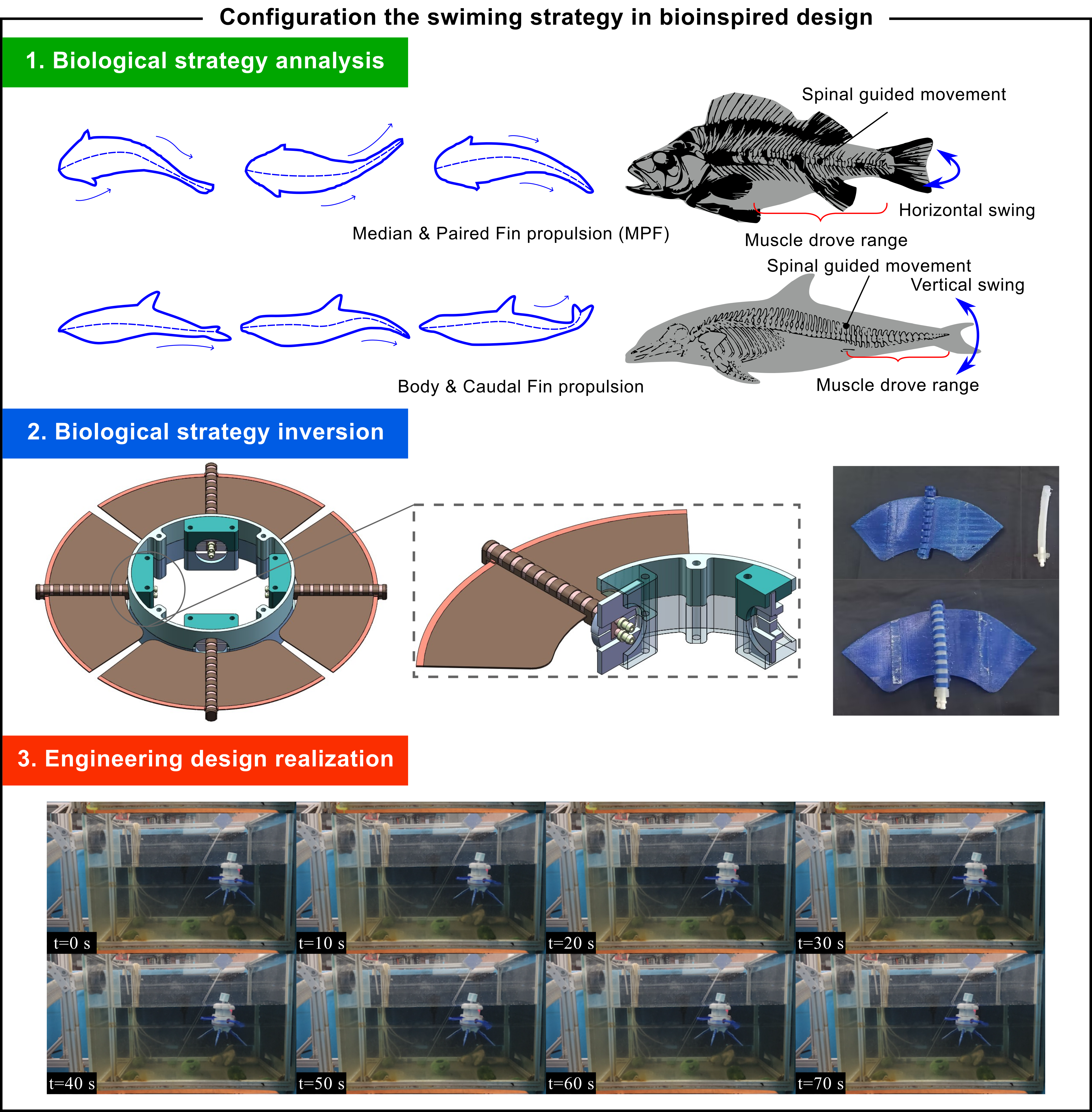}\\
  \caption{Case: tail swing propulsion strategy.}
  \label{fig_case1}
\end{figure*}

As shown in Fig. \ref{fig_case1}, the core difference between fish and whale propulsion strategies lies in the energy transfer path: the former relies on a wave mechanism involving coordinated muscle activity throughout the body; the latter achieves high-speed cruising through oscillations of local flexible structures. This study prioritizes the up-and-down fin flapping pattern as the driving strategy, which is more feasible from an engineering perspective and better meets the design requirements for soft robots.

The mechanical transmission characteristics of the fish's muscle-spinal system provide a biological prototype for soft robot drive design. Experimental studies have shown that the power output from tail oscillation originates from the periodic contraction of red muscle fibers. This biomechanical property can be biomimetically replicated using the driver described earlier. Therefore, the inverse topological structure of the biological strategy based on body/tail fin propulsion mode is shown in Eg. \ref{eq1}.

\begin{equation}\label{eq1}
  S_{\text{base}}^{\text{swim}}=\varGamma \left( S_{e}^{1}\oplus S_{e}^{7}\oplus S_{e}^{9} \right) =\left[ \begin{array}{l}
    \begin{matrix}
    B&		\text{Trust Vector Control}\\
  \end{matrix}\\
    \begin{matrix}
    F&		\left[ \begin{array}{l}
    \text{Driving flexible structure}\\
    \text{The internal structure}\\
    \text{Propagating pressure wave}\\
    \text{Generate directional flow}\\
    \text{Differential steering}\\
  \end{array} \right]\\
  \end{matrix}\\
    \begin{matrix}
    C&		\left[ \begin{array}{l}
    \text{Flexible accessory structure}\\
    \text{Hydrodynamic surface}\\
    \text{Modular tap structure}\\
  \end{array} \right]\\
  \end{matrix}\\
  \end{array} \right] 
\end{equation}

In the manufacturing process of wearable exoskeleton structure, the 3D printing technology of exoskeleton-type driver mentioned above is borrowed to realize the integrated molding of driver and wearable exoskeleton through 3D printing. Fig. \ref{fig_case1} shows the structure of the exoskeleton-guided soft drive with functional structure.

The mobile module is designed by borrowing the structure of biological tail. In order to simplify the control, the movement in all directions is divided into four arc-shaped soft fans. The bidirectional bending mobile module compatible with the unshackled control system is driven by the bionic tail driver, and its structural design is shown in Fig. \ref{fig_case1}

The three-dimensional model of the unmoored underwater soft robot is shown on the left side of Figure 6.8. From top to bottom, it consists of a manually adjustable buoyancy bladder module, a control system in a sealed compartment, a hydraulically driven bending movement module, and other external functional modules. The hardware structure is consistent with the mobile module components of the bionic spider leg pressure regulation system mentioned earlier, adhering to the principle of static reconfigurable models. Each module is connected to the hardware structure using slots, bolts, and other fasteners. The bladder is connected to the hardware structure through an interference fit, utilizing silicone's elastic self-sealing properties and reinforced with nylon thread. The robot communicates wirelessly with a host computer, enabling signal command transmission and firmware upgrades in the air.

During the experiment, the operator sends commands to the soft robot via a host computer to control the working state of the corresponding soft actuators. To verify the mobility of the tetherless soft robot, relevant experiments were conducted in a simulated underwater environment. During movement, the right actuator deforms under pressure, causing it to bend and subsequently drive the connected wearable exoskeleton to bend as well, providing thrust for the robot's forward motion in water. Fig. \ref{fig_case1} shows the robot's movement in water.

From an overall perspective of the movement process, thanks to the centrally symmetric design of the soft robot, it can maintain good balance while moving underwater. The oscillation of its actuators exhibits clear periodic characteristics, which drives the soft robot to move forward at a relatively uniform speed and stability. After preliminary debugging and optimization of the upper-level software, tests on the movement speed of the tetherless soft robot in a simulated underwater environment showed that its average movement speed reached 1.42 mm/s.

\subsection{Soft uncertaintynderwater robot based on fluid injection propulsion strategy}
The jet propulsion system of marine mollusks exhibits unique biomechanical optimization characteristics. Taking cephalopods as an example, their locomotive organs consist of a highly specialized mantle, annular muscle groups, and directional nozzles, forming a composite functional unit. The mantle acts as an elastic energy storage structure in the form of a pressure vessel, with collagen fibers arranged in a helical layer that can store elastic potential energy when relaxed. When the annular muscle groups contract in response to neural signals, the volume of this chamber can decrease by 63\%-67\% within 80 ms to 120 ms, creating an internal pressure gradient of up to 40 kPa. The peak fluid kinetic energy generated by this rapid compression process far exceeds the efficiency level of traditional mechanical pumps[181].

\begin{figure*}[htbp]
  \centering
  \includegraphics[width=12cm]{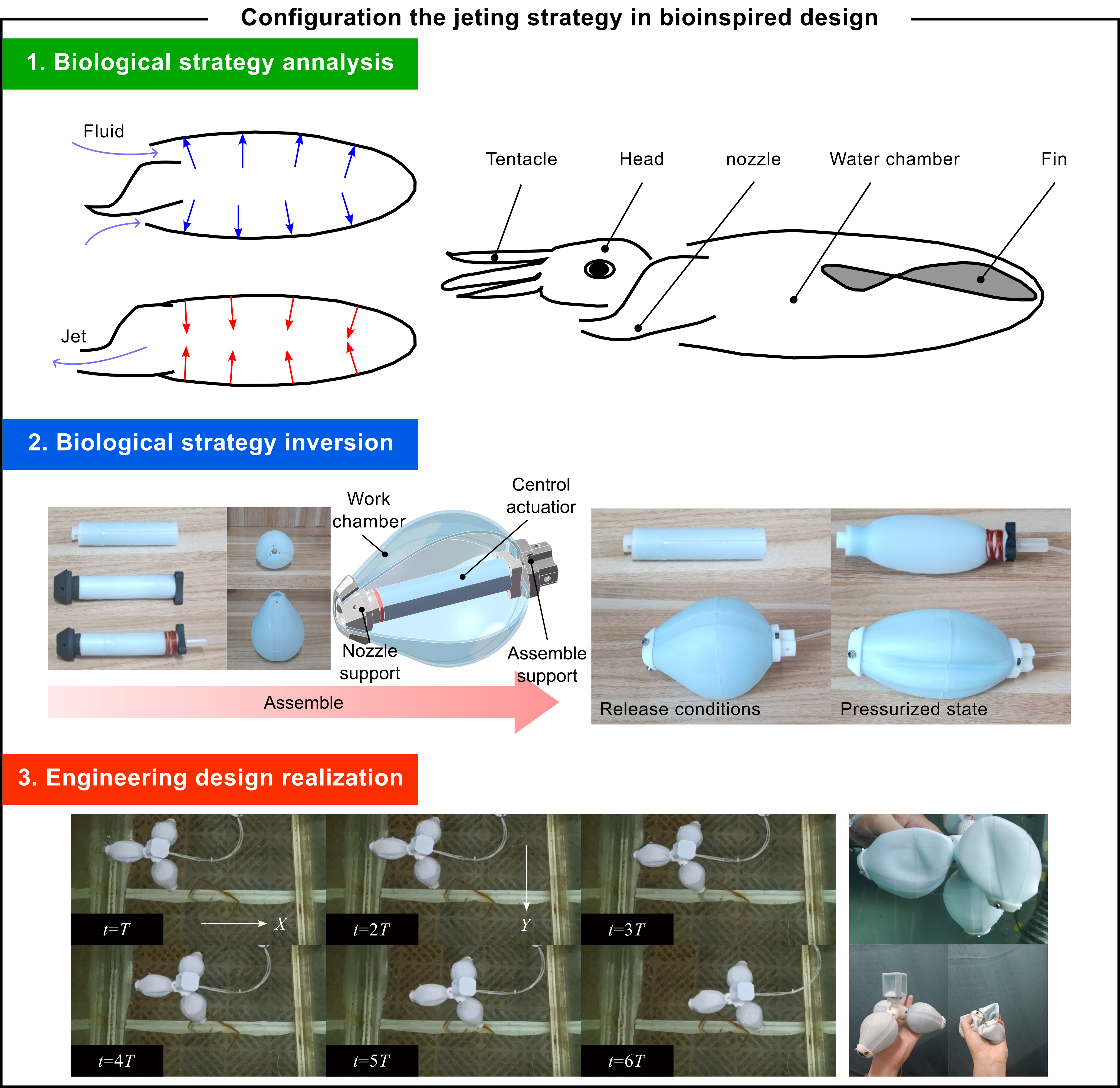}\\
  \caption{Case: fluid injection propulsion strategy.}
  \label{fig_case2}
\end{figure*}

As shown in Fig. \ref{fig_case2}, the morphological adaptability control of the nozzle structure is the core of direction regulation in biological jet systems. The funnel-shaped nozzle of the squid is controlled by two sets of radially arranged skeletal muscles. By changing the cross-sectional shape and jet angle at the outlet, thrust vector regulation can be achieved within a single contraction cycle.

The core functional characteristics of the jet propulsion system in marine organisms can be mapped to soft robotics design through principles of bionics. The elastic energy storage properties of biological pressure chambers can be analogized to pre-stressed balloon structures in robots, both driving fluid ejection through rapid release of elastic potential energy. The active contraction mechanism of biological muscles can be realized by applying pressure to the air sacs of soft actuators. Therefore, the inverse structure of the jet propulsion strategy in marine organisms is shown in Eq. \ref{eq2}.

\begin{equation}\label{eq2}
  S_{\text{base}}^{\text{prop}}=\varGamma \left( S_{e}^{2}\oplus S_{e}^{3} \right) =\left[ \begin{array}{l}
    \begin{matrix}
    B&		\text{Provide underwater thrust}\\
  \end{matrix}\\
    \begin{matrix}
    F&		\left[ \begin{array}{l}
    \text{Rapid fluid removal}\\
    \text{Rotary jet chase}\\
    \text{Recovery by elastic potential energy}\\
    \text{Use external structures to motion}\\
  \end{array} \right]\\
  \end{matrix}\\
    \begin{matrix}
    C&		\left[ \begin{array}{l}
    \text{Drive mechanism extrusion}\\
    \text{Nozzle based on rigid support}\\
    \text{Exterior diamond-shaped structure}\\
    \text{Elastic cavity }\\
  \end{array} \right]\\
  \end{matrix}\\
  \end{array} \right] 
\end{equation}

Fig. \ref{fig_case2} illustrates the manufacturing process of the jet unit. Its soft sections include an airbag structure for the central driver and a working bladder for fluid entry and exit. All soft parts of the robot are formed through silicone curing. Silicone with a Shore hardness of 36A (Smooth-ON SIL 936, smoothOn) is selected as the soft material. The mold is made using 3D printing. Throughout the process, a vacuum pump is used to remove bubbles from the liquid silicone. Figure shows the manufacturing process of the sealed connection between the tubular body and the drive bladder. The driving principle of the central driver is similar to that of the elongated driver in the fiber-embedded driver mentioned earlier. To connect the nozzle, Rossing et al.'s method proposed in [182] for connecting rigid parts to silicone is adopted. The central driver drives the entire working bladder. According to the figure, the drive bladder is assembled to form the central driver.

This paper uses the pressure control platform shown in Fig. \ref{fig_case2} to provide high-pressure gas and water. The pump pressurizes the fluid and inputs it into the valve group for pressure regulation and distribution. In the experiment, a maximum working pressure of 100 KPa was selected. The pressure detected during the experiment is provided by a pressure transmitter (MEACON, model MIK-P300). Mechanical experiments include measuring the force output and flow rate output of the center driver and unit. A HANDPI (HLD0824) force gauge is used to measure the fluid mass and force generated by the actuator. The thrust of the injection unit is measured using a HANDPI (HLD2000) force gauge. The sampling frequency of the force gauges is 10 Hz, with an accuracy of 0.001 N. The sampling frequency of the pressure transmitter is also 10 Hz, with an accuracy of 0.1 KPa.

\subsection{Soft underwater crawling robot based on autonomic peristalsis}
\begin{figure*}[htbp]
  \centering
  \includegraphics[width=12cm]{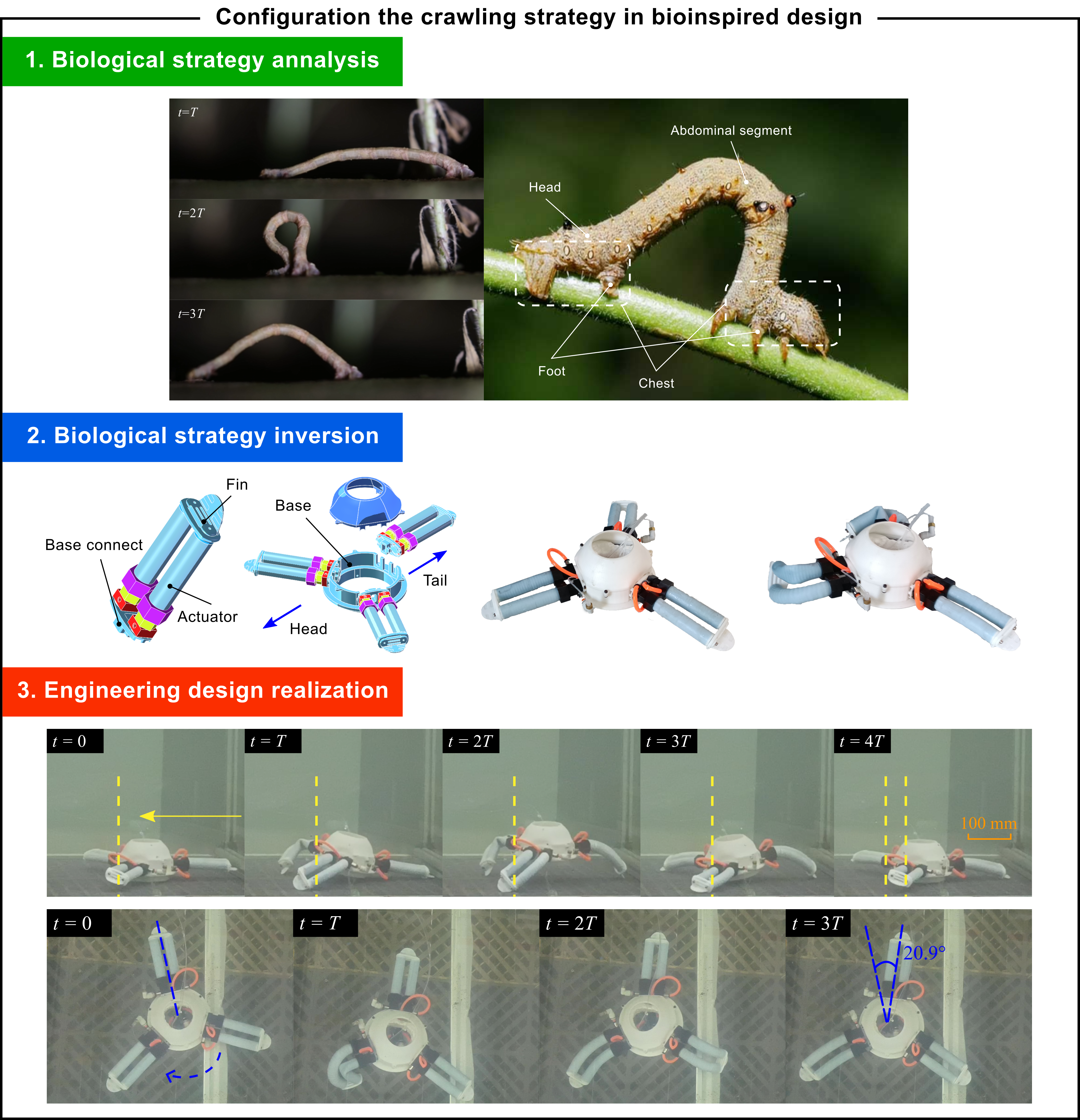}\\
  \caption{Case: crawling strategy based on autonomic peristalsis.}
  \label{fig_case3}
\end{figure*}

As show in Fig. \ref{fig_case3}, the displacement increment was generated, and the stride length could reach 68.5±7.2\% of the body length. At the same time, the pseudopod of the tail separated from the base to complete the movement cycle.

The static water skeleton of the inchworm is the core structure that enables its anisotropic deformation. Its epidermis consists of collagen fiber layers arranged orthogonally, with the outer cuticle having a significantly higher Young's modulus than the inner epidermis, creating a gradient stiffness distribution. Muscle tissue is embedded at a 45° interdigitated angle in the basal layer of the epidermis, forming a three-layer orthogonal fiber network. The longitudinal muscle fibers have a tensile strength of 12.7 MPa ± 2.1 MPa, while the circular muscle fibers enhance contraction efficiency through strain energy storage mechanisms. This structural characteristic allows the circular muscles to passively expand under the Poisson effect during longitudinal muscle contraction, converting hydraulic energy into axial thrust.

In biomimetic design, the driving array can be used to simulate the pattern of muscle movement distribution, so as to construct the crawling motion strategy of soft robots. The inverse structure from biological strategy to engineering application is shown in Equation (6 5).

\begin{equation}\label{eq3}
  S_{\text{base}}^{\text{cra}}=\varGamma \left( S_{e}^{4}\oplus S_{e}^{5} \right) =\left[ \begin{array}{l}
    \begin{matrix}
    B&		\text{Achieve crawling}\\
  \end{matrix}\\
    \begin{matrix}
    F&		\left[ \begin{array}{l}
    \text{Coordinate muscle movement}\\
    \text{Shrink driver}\\
    \text{Release the drive from the base}\\
    \text{Transmitted contraction wave}\\
  \end{array} \right]\\
  \end{matrix}\\
    \begin{matrix}
    C&		\left[ \begin{array}{c}
    \text{Radial symmetric structure}\\
    \text{Contact friction control}\\
    \text{Still water skeleton system}\\
  \end{array} \right]\\
  \end{matrix}\\
  \end{array} \right] 
\end{equation}

According to the assembly method shown in Fig. \ref{fig_case3}, the drive module is assembled from actuators, with rigid fins designed to propel the robot underwater or on the seafloor. This soft robot is equipped with two drive modules. The drive module at the head needs to have three-dimensional bending capability, while the drive module at the tail only requires downward bending. Apart from the arrangement of the actuators within the drive modules, both types of actuators look identical. The layout of the actuators in the drive modules is shown in Fig. \ref{fig_case3}, enabling the robot to achieve the behavioral effects depicted in Fig. \ref{fig_case3}. The connection between the actuators and the drive modules is a hinge. By determining the end coordinates of the actuators in the drive modules, the position of the rigid fins in water and their output force can be obtained. According to the layout shown in Fig. \ref{fig_case3}, the drive module is fixed to the base through its connections, as illustrated in Fig. \ref{fig_case3}.

Consider the two drive modules at the head as a link on the centerline, ensuring the robot's movement aligns with the crawling model shown in Fig. \ref{fig_case3}. Due to the low friction coefficient, the robot slides along the bottom. Nevertheless, the robot still completes a rotation. In experiments, the robot can crawl forward 46 mm or rotate 20.9° within one cycle of pressure, with no relative sliding between the contact point and the bottom. The pressure cycle depends on the size of the chamber, input flux, and degree of deformation. Experimental conditions are shown in Fig. \ref{fig_case3}.


\section{Conclusion}
This work establishes an F-B-C-E bio-inspired design framework synergizing large language models (LLMs) and knowledge graphs for systematic biological strategy screening. Key advancements include: Bio-inspired strategy mapping: A method integrating LLMs and knowledge graphs under the Function-Behavior-Characteristic-Environment (F-B-C-E) framework enables rapid biological prototype screening. LLMs extract engineering requirements and construct conceptual keyword networks via knowledge graphs, while systematic biological prototype analysis across four dimensions (function, behavior, characteristic, environment) bridges biological logic with engineering analogies. A hybrid multi-criteria decision-making (MCDM) model optimizes engineering strategies using six indicators (functional compliance, behavioral alignment, characteristic consistency, environmental migration potential, reliability, and economic tolerance).

A fiber-guided soft actuator achieving 37\% motion enhancement via circumferential-radial fiber synergy, validated in wearable exoskeletons;
Spider-leg-inspired hydraulic bridges (Type A: 0-100\% linear control; Type C: <5\% fluctuation at 0.6-1.2 MPa) and biomimetic interfaces (<1.2 s docking, <8\% pressure loss);
A multimodal underwater propulsion system integrating tail-fin (1.42 mm/s), squid-jet (24 mm/s at 100 kPa), inchworm (46 mm/cycle), and jellyfish strategies, achieving 18-32\% energy efficiency gains.
The framework demonstrates cross-scale applicability from component design to system-level pressure regulation and motion coordination, validated through rigorous biomechanical experiments. Hybrid decision-making models address strategy optimization under multi-constraint scenarios, while gradient structural biomimetics enables functional evolution of soft actuators.

Future research will focus on: 1) Knowledge-enhanced strategy inversion via thrust modeling and dynamic similarity analysis; 2) High-fidelity biological prototype characterization using micro-CT and fluid-structure interaction simulations; 3) Cross-species motion mechanism decoding to unify propulsion metrics. These efforts aim to establish quantifiable bio-to-engineering translation protocols for adaptive robotics.

\section{Acknowledgement}
This work was supported by the National Key Research and Development Program of China 2023YFB4704604-03, Natural Science Foundation of China under Grant 51875113, Natural Science Joint Guidance Foundation of the Heilongjiang Province of China under Grant LH2019E027, Opening Project of the Key Laboratory of Bionic Engineering (Ministry of Education) by Jilin University.


\bibliography{mybibfile}

\begin{thebibliography}{10}
\expandafter\ifx\csname url\endcsname\relax
  \def\url#1{\texttt{#1}}\fi
\expandafter\ifx\csname urlprefix\endcsname\relax\def\urlprefix{URL }\fi
\expandafter\ifx\csname href\endcsname\relax
  \def\href#1#2{#2} \def\path#1{#1}\fi

\bibitem{RN259}
S.~Feng, X.~He, \href{http://dx.doi.org/10.1007/s42242-020-00079-3}{A review of biomimetic research for erosion wear resistance}, Bio-Design and Manufacturing 3~(4) (2020) 331--347.
\newblock \href {http://dx.doi.org/10.1007/s42242-020-00079-3} {\path{doi:10.1007/s42242-020-00079-3}}.
\newline\urlprefix\url{http://dx.doi.org/10.1007/s42242-020-00079-3}

\bibitem{RN260}
G.~Alessio, R.~Zoë, B.~Vincent, \href{http://dx.doi.org/10.1007/s13347-023-00665-0}{What does it mean to mimic nature? a typology for biomimetic design}, Philosophy \&amp; Technology 36~(4).
\newblock \href {http://dx.doi.org/10.1007/s13347-023-00665-0} {\path{doi:10.1007/s13347-023-00665-0}}.
\newline\urlprefix\url{http://dx.doi.org/10.1007/s13347-023-00665-0}

\bibitem{RN261}
K.~Sun-Joong, L.~Ji-Hyun, \href{http://dx.doi.org/10.1016/j.engappai.2016.10.003}{A study on metadata structure and recommenders of biological systems to support bio-inspired design}, Engineering Applications of Artificial Intelligence 57 (2017) 16--41.
\newblock \href {http://dx.doi.org/10.1016/j.engappai.2016.10.003} {\path{doi:10.1016/j.engappai.2016.10.003}}.
\newline\urlprefix\url{http://dx.doi.org/10.1016/j.engappai.2016.10.003}

\bibitem{RN262}
Y.~Gülşen~Töre, F.~Roxana~Moroşanu, C.~Nathan, \href{http://dx.doi.org/10.1016/j.destud.2017.11.006}{User requirements for analogical design support tools: Learning from practitioners of bio-inspired design}, Design Studies 58 (2018) 1--35.
\newblock \href {http://dx.doi.org/10.1016/j.destud.2017.11.006} {\path{doi:10.1016/j.destud.2017.11.006}}.
\newline\urlprefix\url{http://dx.doi.org/10.1016/j.destud.2017.11.006}

\bibitem{RN263}
J.~M. Sarah, K.~Banafsheh, M.~G. Austin, H.~Thibaut, K.~U. Colleen, R.~Ariana, W.~Nicholas, F.~V.~V. Julian, K.~S.~N. Jacquelyn, H.~N. Peter, \href{http://dx.doi.org/10.3390/designs2040053}{E2bmo: Facilitating user interaction with a biomimetic ontology via semantic translation and interface design}, Designs 2~(4) (2018) 53.
\newblock \href {http://dx.doi.org/10.3390/designs2040053} {\path{doi:10.3390/designs2040053}}.
\newline\urlprefix\url{http://dx.doi.org/10.3390/designs2040053}

\bibitem{RN264}
L.~Linli, G.~Fu, L.~Hao, G.~Jianfeng, G.~Xinjian, \href{http://dx.doi.org/10.1109/access.2021.3108218}{Digital twin bionics: A biological evolution-based digital twin approach for rapid product development}, IEEE Access 9 (2021) 121507--121521.
\newblock \href {http://dx.doi.org/10.1109/access.2021.3108218} {\path{doi:10.1109/access.2021.3108218}}.
\newline\urlprefix\url{http://dx.doi.org/10.1109/access.2021.3108218}

\bibitem{RN265}
K.~S.~N. Jacquelyn, L.~N. Robert, E.~Marjan, \href{http://dx.doi.org/10.1115/detc2013-12068}{Teaching biomimicry with an engineering-to-biology thesaurus}, Volume 1: 15th International Conference on Advanced Vehicle Technologies; 10th International Conference on Design Education; 7th International Conference on Micro- and Nanosystems\href {http://dx.doi.org/10.1115/detc2013-12068} {\path{doi:10.1115/detc2013-12068}}.
\newline\urlprefix\url{http://dx.doi.org/10.1115/detc2013-12068}

\bibitem{RN266}
K.~G. Ashok, H.~William, \href{http://dx.doi.org/10.1017/dsj.2021.23}{Multifunctional and domain independent? a meta-analysis of case studies of biologically inspired design}, Design Science 7.
\newblock \href {http://dx.doi.org/10.1017/dsj.2021.23} {\path{doi:10.1017/dsj.2021.23}}.
\newline\urlprefix\url{http://dx.doi.org/10.1017/dsj.2021.23}

\bibitem{RN267}
A.~Negar, R.~Jean, \href{http://dx.doi.org/10.1016/j.ins.2012.04.033}{An application of multi-criteria decision aids models for case-based reasoning}, Information Sciences 210 (2012) 55--66.
\newblock \href {http://dx.doi.org/10.1016/j.ins.2012.04.033} {\path{doi:10.1016/j.ins.2012.04.033}}.
\newline\urlprefix\url{http://dx.doi.org/10.1016/j.ins.2012.04.033}

\bibitem{RN268}
B.~Zhonghang, M.~Lei, L.~Hsiung-Cheng, \href{http://dx.doi.org/10.3390/su12104276}{Green product design based on the biotriz multi-contradiction resolution method}, Sustainability 12~(10) (2020) 4276.
\newblock \href {http://dx.doi.org/10.3390/su12104276} {\path{doi:10.3390/su12104276}}.
\newline\urlprefix\url{http://dx.doi.org/10.3390/su12104276}

\bibitem{RN269}
Z.~Mingzhi, C.~Huijing, Z.~Mingming, L.~Caicai, Z.~Teng, \href{http://dx.doi.org/10.1088/1748-3190/ac03bd}{Optimal design of aeroacoustic airfoils with owl-inspired trailing-edge serrations}, Bioinspiration \&amp; Biomimetics 16~(5) (2021) 56004.
\newblock \href {http://dx.doi.org/10.1088/1748-3190/ac03bd} {\path{doi:10.1088/1748-3190/ac03bd}}.
\newline\urlprefix\url{http://dx.doi.org/10.1088/1748-3190/ac03bd}

\bibitem{RN270}
Z.~Yong, S.~Xiang, Z.~Xufeng, X.~Xiaojie, W.~Jun, C.~Peining, S.~Xuemei, P.~Huisheng, \href{http://dx.doi.org/10.1002/adfm.202005200}{Flexible color‐tunable electroluminescent devices by designing dielectric‐distinguishing double‐stacked emissive layers}, Advanced Functional Materials 30~(50).
\newblock \href {http://dx.doi.org/10.1002/adfm.202005200} {\path{doi:10.1002/adfm.202005200}}.
\newline\urlprefix\url{http://dx.doi.org/10.1002/adfm.202005200}

\end{thebibliography}

\end{document}